\documentclass[runningheads]{llncs}

\usepackage[final,year=2026]{eccv}

\usepackage{eccvabbrv}

\usepackage{graphicx}
\usepackage{booktabs}

\usepackage[dvipsnames]{xcolor}
\usepackage{pifont}
\newcommand{\gou}{{\color{OliveGreen}\ding{51}}}
\newcommand{\cha}{{\color{BrickRed}\ding{55}}}
\usepackage[utf8]{inputenc}
\usepackage[T1]{fontenc}
\usepackage{url}
\usepackage{amsfonts}
\usepackage{nicefrac}
\usepackage{microtype}
\usepackage{amsmath,amsfonts}
\usepackage[ruled,vlined]{algorithm2e}
\usepackage{array}
\usepackage{textcomp}
\usepackage{stfloats}
\usepackage{wrapfig}
\usepackage{verbatim}
\usepackage[normalem]{ulem}

\usepackage{multirow}
\usepackage{multicol} 
\usepackage{float}             
\usepackage{makecell}
\usepackage{amsmath}
\usepackage{amsmath,amssymb}
\usepackage{mathrsfs}
\usepackage{tcolorbox}
\usepackage{xspace}
\usepackage{enumitem}
\usepackage{xcolor}
\usepackage{colortbl}

\usepackage{algpseudocode}
\definecolor{backcolor}{RGB}{232, 242, 255}
\usepackage{fontawesome5} 
\usepackage[framemethod=TikZ]{mdframed}

\usepackage[breaklinks,colorlinks,citecolor=eccvblue]{hyperref}

\usepackage{orcidlink}

\begin{document}

\title{Towards Interactive Global Geolocation Assistant} 

\author{
Zhiyang Dou\inst{1,\star} \and
Zipeng Wang\inst{1,\star} \and
Xumeng Han\inst{2,\star} \\
Guorong Li\inst{3} \and
Zhenjun Han\inst{2,\dagger} \and
Zhipei Huang\inst{2}
}

\authorrunning{Z.~Dou et al.}

\institute{
School of Advanced Interdisciplinary Sciences, University of Chinese Academy of Sciences (UCAS), Beijing, China
\and
School of Electronic, Electrical and Communication Engineering, UCAS, Beijing, China
\and
School of Computer Science and Technology, UCAS, Beijing, China\\
\email{\faEnvelope\, hanzhj@ucas.ac.cn}\\
$^\star$ Equal contribution. \quad $^\dagger$ Corresponding author.
}

\maketitle

\begin{abstract}
Global geolocation, the task of predicting precise coordinates from street-view imagery, is inherently plagued by visual ambiguity. Resolving such ambiguity necessitates a transition from static, one-shot predictions to an \textbf{interactive geolocation paradigm} driven by multi-turn deductive reasoning. However, most of existing geolocation models and general-purpose MLLMs fail to support this dynamic process, mainly due to the lack of interaction capabilities and the geographic knowledge gap. Driven by the imperative to actualize this interactive paradigm, we introduce \textbf{MG-Geo}, the first large-scale multimodal geolocation dataset explicitly structured for spatial reasoning. Comprising 4.87M geo-tagged Meta entries, 70K image-grounded Clue samples, and 73K multi-turn Dialog samples across 210 countries and territories, MG-Geo separates large-scale geographic alignment from reasoning-oriented supervision. Experiments demonstrate that GaGA achieves SOTA performance across several benchmarks. Notably, on the GWS15k dataset, it surpasses the strong Hybrid Model by 4.57\% and 2.92\% at the country and city levels, respectively, while securing the highest city-level accuracy (7.46\%) on OSV-5M-test. More importantly, we formalize the ``Similarity Trap''—a phenomenon where distributive visual features mislead static models—and demonstrate that GaGA effectively navigates this challenge through a \textit{Tiered Interaction Protocol}. By dynamically integrating user-provided geographic anchors, GaGA achieves significant localization improvements. Our dataset is accessible via: \url{https://huggingface.co/datasets/kendouvg/MG-Geo}.

\end{abstract}

\section{Introduction}
Global geolocation aims to predict the exact location of any street-view image, with wide applications in security surveillance, emergency response, disease outbreak prediction, environmental monitoring, and tourism navigation~\cite{metageo, gpt4geo, geollm}. This task requires integrating visual cues, such as road signs, architectural styles, climate, and vegetation, with geographic knowledge to accurately predict GPS coordinates or location labels. For images with landmarks or distinctive architecture, the location can be inferred by combining visual features with contextual knowledge. However, geolocation becomes more challenging in homogenous environments, such as highways or natural landscapes, where subtle geographic clues like road markings, license plate types, and signage must be relied upon.

Traditional street-view localization methods are generally categorized into retrieval-based and classification-based approaches. The retrieval-based methods~\cite{TransGeo,cross_view_loc,zhang2023cross,GeoCLIP} match input images with similar ones from a geotagged database but are constrained by the diversity and completeness of the database. The classification-based methods~\cite{weyand2016planet,seo2018,Translocator} classify images into predefined regions based on visual features, but they lack interpretability and fail to provide explicit visual cues. In practical scenarios, geolocation is rarely a one-time, static process; it involves integrating and refining multiple sources of information iteratively through continuous interaction. Traditional geolocation models inherently lack interpretability and flexibility.

Multimodal Large Language Models (MLLMs)~\cite{llava, internvl, llama3} offer a promising alternative due to their ability to synthesize cross-modal information and perform complex interpretive reasoning. This reasoning capacity is the key to unlocking an \textbf{interactive geolocation paradigm}, where spatial ambiguities can be dynamically resolved through multi-turn dialogue, hypothesis generation, and evidence accumulation. Yet, existing general-purpose MLLMs struggle to execute this interactive process, primarily due to a profound geographic knowledge gap~\cite{gpt4geo}. They fail to establish robust, verifiable associations between granular geographic elements (e.g., specific road markings or vegetation patterns) and precise spatial coordinates. 

\begin{figure}[!htbp]
\centering
\includegraphics[width=1.0\columnwidth]{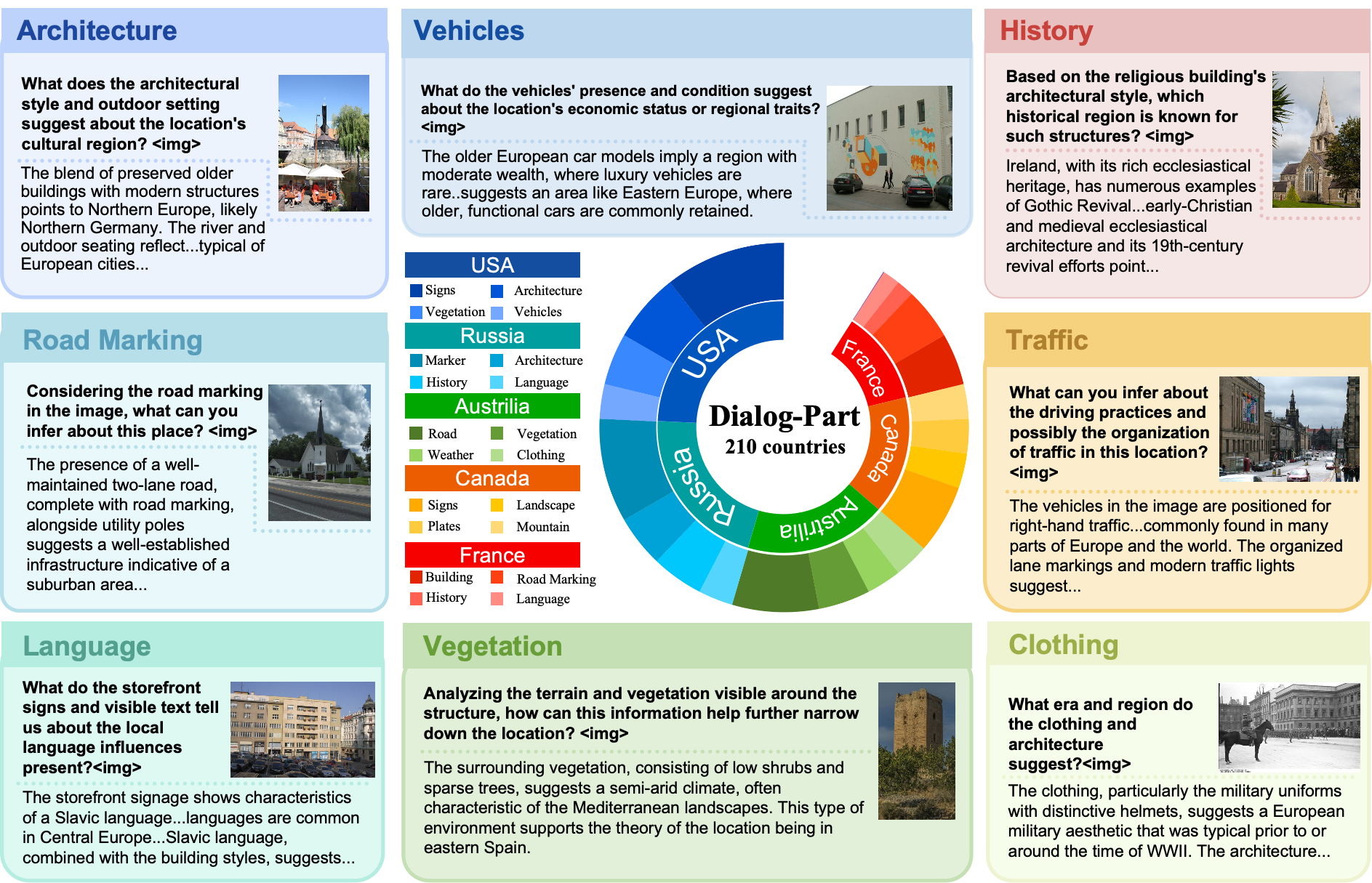} 
\caption{Illustration of MG-Geo. Featuring diverse geographic scenes and visual cues, these images demonstrate the utility for training MLLMs to connect visual content with geographic locations and enrich their understanding of global environments.}
\label{fig:Fig1}
\end{figure}

A major bottleneck preventing this paradigm shift is the fundamental misalignment of existing training data. While conventional geolocation datasets, such as OSV-5M~\cite{OSV5M}, are engineered exclusively for static, one-shot mapping, general multimodal datasets~\cite{sharegpt, datacomp} lack the necessary geographic granularity and administrative hierarchies. Driven by the imperative to actualize this interactive geolocation paradigm, we introduce \textbf{MG-Geo}, the first large-scale multimodal dataset explicitly structured to support multi-turn dialogue and deductive spatial reasoning. Featuring an \textit{Interactive Reasoning Chain-of-Thought} strategy, MG-Geo contains 4.87M geo-tagged Meta entries, 70K image-grounded Clue samples, and 73K multi-turn Dialog samples, with the Clue and Dialog parts organized around eight categories of Geographic Element Cues. As depicted in Figure~\ref{fig:Fig1}, it covers 210 countries and categorizes cues into eight critical dimensions—including road markings, vegetation, and linguistic signatures. By transitioning from isolated static labels to interconnected reasoning chains, MG-Geo provides the essential foundation for training MLLMs to perform dynamic, evidence-based geographic assistance.

Utilizing this resource, we propose the \textbf{G}lob\textbf{a}l \textbf{G}eolocation \textbf{A}ssistant (GaGA), an MLLM architected for high-precision, interactive spatial reasoning. GaGA excels not only in static benchmarks but also in navigating the \textbf{``Similarity Trap''}—a fundamental challenge where geographically widespread visual motifs (e.g., neoclassical architecture) lead to over-confident misclassifications. Our contributions are summarized as follows:

(1) We present \textbf{MG-Geo}, a large-scale multimodal geolocation dataset with 4.87M geo-tagged Meta entries, 70K image-grounded Clue samples, and 73K multi-turn Dialog samples, enabling MLLMs to move beyond static image tagging toward geographic reasoning and interaction.

(2) We develop \textbf{GaGA}, which establishes a new state-of-the-art (SOTA) across multiple benchmarks. On the GWS15k dataset, GaGA surpasses the strong Hybrid baseline by 4.57\% and 2.92\% at the country and city levels, respectively.

(3) We formalize and investigate the \textbf{Interactive Geolocation Paradigm}. Through a \textbf{Tiered Interaction Protocol}, we demonstrate how GaGA can effectively leverage user-provided anchors to resolve visual ambiguity.

\section{Related work}

\subsection{Geolocation Datasets}
In the domain of geolocation, the localizability of images within datasets is of paramount importance. Although the existing datasets, such as Im2GPS3k~\cite{vo2017revisiting}, YFCC4K~\cite{yfcc100m} and MP-16~\cite{theiner2022interpretable} contain a large number of geotagged images, many of them are difficult to localize and exhibit distribution biases. GWS15k~\cite{gws15k} mitigates distribution differences, and ensures that the images are authentic, localizable street views. OSV-5M~\cite{OSV5M} is the largest open-source collection of planet-scale, localizable street view images. The Google Landmark V2~\cite{GoogleLandmark} dataset contains globally distributed human-made and natural landmarks and showcase iconic landscapes.

We introduce MG-Geo, the first large-scale multimodal geolocation dataset comprising 4.87M geo-tagged Meta entries, 70K image-grounded Clue samples, and 73K multi-turn Dialog samples structured with an expert-guided Chain-of-Thought strategy to enhance the geospatial reasoning and interactivity of MLLMs. MG-Geo improves geographic coverage over existing reasoning datasets by combining OSV-5M street-view metadata with GLDv2 landmark imagery, while we visualize its global coordinate distribution and long-tail nature in the Appendix~\ref{app:geo_distribution}. It also incorporates well-structured global language knowledge with eight categories of detailed geographic element cues, providing a dataset that better reflects the complexity and diversity of real-world geolocation challenges.
\begin{table}[!t]
\centering
\caption{\textbf{Comparison between MG-Geo and existing datasets}. 
MG-Geo is the first large-scale multimodal dataset featuring \textbf{expert-guided CoT}, \textbf{8-category geographic clues}, and \textbf{multi-turn interactive dialogues}. 
}
\resizebox{\linewidth}{!}{
  \begin{tabular}{@{}l|c|c|c|c|c|c|c|c@{}}
    \specialrule{\heavyrulewidth}{0pt}{0pt}
    \multirow{2}{*}{Dataset} & \multirow{2}{*}{Size} & \multirow{2}{*}{Open} & \multirow{2}{*}{Source} & \multirow{2}{*}{Scope} & \multicolumn{2}{c|}{Chain of Thought} & \multirow{2}{*}{Geo-Clues} & \multicolumn{1}{c}{Multi-turn} \\
    & & & & & AI-gen & Expert-guided & & Dialog \\
    \specialrule{\lightrulewidth}{0pt}{0pt}
    MP-16\cite{theiner2022interpretable} & 4.7M & \gou & Web & Biased & \cha & \cha & \cha & \cha \\
    OSV-5M\cite{OSV5M} & 5M & \gou & Street-view & Global & \cha & \cha & \cha & \cha \\
    Google Landmark V2\cite{GoogleLandmark} & 5M & \gou & Landmark & Global & \cha & \cha & \cha & \cha \\
    \specialrule{\lightrulewidth}{0pt}{0pt}
    MP16-Reason~\cite{GLOBE} & 33k & \gou & Web & Biased & \gou & \cha & \gou & \cha \\
    GRE30k~\cite{gre} & 30k & \gou & Web & Biased & \gou & \cha & \cha & \cha \\
    \rowcolor{backcolor}
    \textbf{MG-Geo (ours)} & \textbf{5M} & \gou & \makecell{\textbf{Street-view} \\ + \textbf{Landmark}} & \textbf{Global} & \gou & \textbf{\gou} & \textbf{\gou} & \textbf{\gou} \\
    \specialrule{\heavyrulewidth}{0pt}{0pt}  \end{tabular}
}\label{tab:dataset_comparison}
\end{table}

\subsection{Traditional Geolocation Methods}
Traditional mainstream geolocation methods can be broadly categorized into two approaches: image-based retrieval and classification-based methods. Image-to-image retrieval techniques rely on dense image retrieval libraries, which perform well for localization tasks within small areas. However, the cost of constructing such retrieval libraries on a global scale is prohibitively high. When geolocation is treated as a classification task, categories can be defined based on administrative regions, divided into geocells according to specific rules, or discretized into latitude and longitude coordinates. TransLocator~\cite{Translocator} employs images and semantic segmentation maps as inputs, facilitating interaction between two parallel branches after each Transformer layer and enabling multitask geolocation and scene recognition. GeoCLIP~\cite{GeoCLIP} introduces a location encoder and applies random Fourier feature representations to latitude and longitude coordinates. It utilizes the pretrained CLIP~\cite{CLIP} visual encoder to represent images and aligns them with the corresponding location features for localization. PIGEON~\cite{Pigeon} classifies within a self-created geocell and retrieves locations within clusters.

\subsection{Generation-Based Geolocation Methods}
Powerful MLLMs have injected new momentum into geolocation research. We refer to this emerging paradigm as generation-based geolocation, which has given rise to two predominant research directions. One dominant approach combines MLLMs with Retrieval-Augmented Generation (RAG) to enhance precision. This line of work includes Img2Loc~\cite{Img2Loc}, which reframes geolocation as a training-free text generation problem using CLIP-based retrieval; G3~\cite{G3} proposes a three-stage framework to mitigate challenges of visual semantics and data imbalance; and GeoRanker~\cite{GeoRanker} introduces a distance-aware ranking framework to model spatial relationships among retrieved candidates.
Another research direction focuses on enhancing the intrinsic reasoning abilities of MLLMs. For example, GeoReasoner~\cite{GeoReasoner} fine-tunes an MLLM for street view localization using a curated dataset of human reasoning patterns derived from geolocation games. Similarly, GLOBE~\cite{GLOBE} and GRE~\cite{gre} develop a reasoning-oriented dataset from diverse social media images and use group-relative policy optimization to improve the generation of visual-clue-based rationales.
Distinguished from prior work, GaGA is trained on a planet-scale dataset and introduces a novel interactive reasoning paradigm via dialogue. 

\section{MG-Geo Dataset}\label{mggeo}
In this paper, we introduce MG-Geo, a novel dataset encompassing a diverse array of geographic element cues, including architecture, environment, landmarks, and climate across various countries. The dataset is structured into three distinct components: the \textit{Meta Part}, the \textit{Clue Part}, and the \textit{Dialog Part}. We leverage structured geographic knowledge, elicited from expert GeoGuessr players, and human-guided interactions with powerful MLLMs to facilitate the construction of this dataset. 

\subsection{Meta Part}
In the Meta Part, images and meta-geographic information are taken from the OSV-5M~\cite{OSV5M}, which inherits its characteristics of good distribution, wide scope, and high quality. After removing a small number of samples with incomplete location annotations, we organize each sample into JSON format using three levels of administrative boundaries—country, region, and city. This results in a total of 4.87 million entries, covering 70k cities, 2.7k regions, and 210 countries.

\begin{figure}[!htbp] 
\centering
\includegraphics[width=1\linewidth]{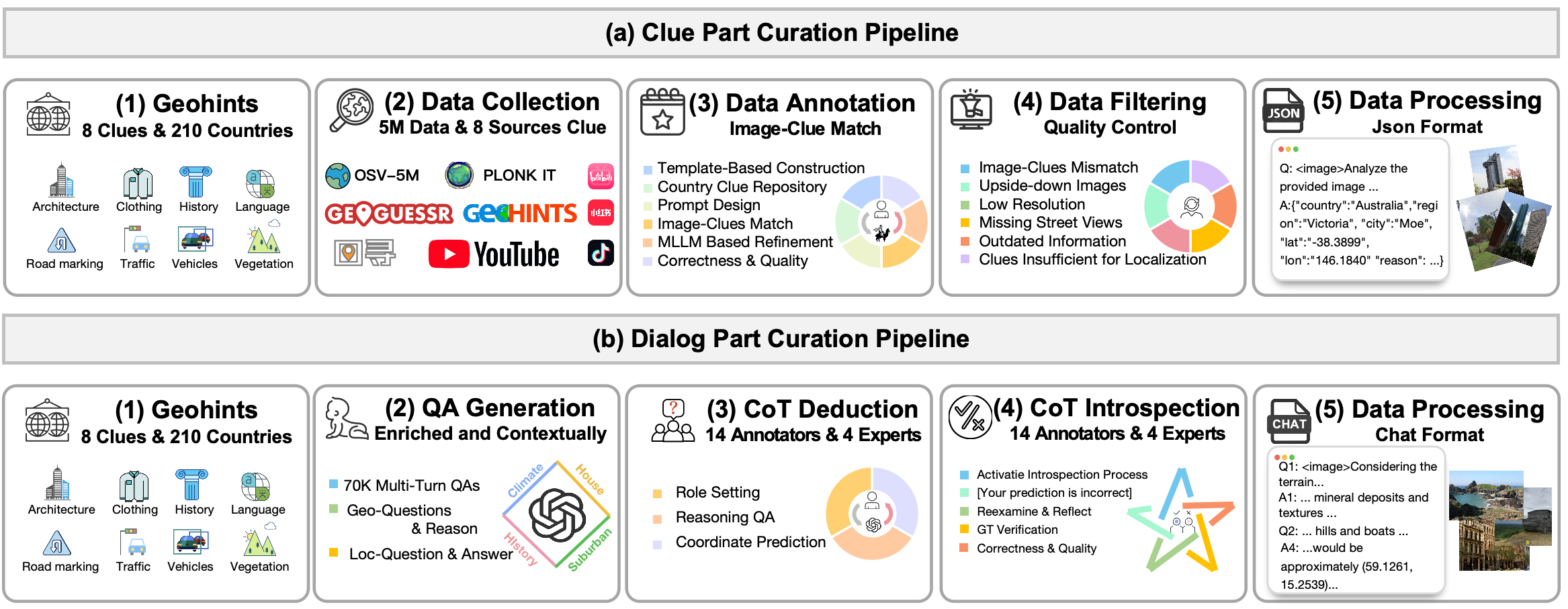}
\caption{\textbf{An illustration of our pipeline for data curation.} (a) We construct the Clue Part by leveraging guidance clues from online geolocation game communities and employing an MLLM. (b) We generate location-agnostic, multi-turn reasoning QA pairs and high-quality dialog data for the Dialog Part, applying the Interactive Reasoning CoT method to activate CoT Deduction and CoT Introspection tasks.}
\label{fig:Data_Statistics}
\end{figure}

\subsection{Clue Part}
To enhance interpretability, we develop an automated multimodal QA generation paradigm that transforms raw geographic annotations into structured image-text clue pairs. Figure~\ref{fig:Data_Statistics}(a) illustrates this high-correlation data generation pipeline.

\textbf{MLLM-Based Refinement.}
While the 3,000 expert clues from GeoGuessr and Tuxun provide rich localization knowledge, pure text lacks the visual context necessary for effective multimodal reasoning. To address this, we leverage MLLMs to align each clue with its corresponding visual features. We sampled 70k globally distributed images from OSV-5M~\cite{OSV5M} and implemented a two-step process: (1) \textit{constructing country-specific clue repositories} via manual classification, and (2) \textit{automated image-clue matching} to associate street-view imagery with relevant regional descriptors.

\textbf{Human Verification.}
Despite rigorous automated matching, visual ambiguities (e.g., low resolution or missing views) may persist. We implemented a manual validation protocol to identify and rectify erroneous pairs. Evaluators trace error sources to either prune problematic samples or recalibrate metadata, ensuring high fidelity between natural language descriptions and their visual targets.

\subsection{Dialog Part}\label{dialog part}
As shown in Figure~\ref{fig:Data_Statistics}(b), we begin the \textit{Dialog Part} construction process by standardizing a well-annotated subset of the Google Landmark V2 into a unified metadata structure, ensuring the generation of multi-turn reasoning QA pairs that are location-agnostic. In order to enhance GaGA’s reasoning depth and conversational ability by supporting the analysis of images from multiple perspectives and inferring specific locations, we select 73K samples from Google Landmark V2 with rich information such as architecture, vegetation, cultural elements, and climate. Then, with the assistance of GPT-4V, we generate QA pairs using the Interactive Reasoning CoT method.

\textbf{Question-Answer Generation}
We intend to create image descriptions that thoroughly capture visible appearance and attributes, integrating relevant knowledge, climatic characteristics, architectural styles, and even historical context. This all-encompassing strategy ensures the dataset's robust support for a broad spectrum of real-world applications by providing enriched and contextually rich data. 

The generation of multi-turn QAs mainly relies on providing unified metadata and carefully designed prompts to MLLMs, specifically GPT-4V. Through this process, GPT-4V engages in multi-turn self-questioning based on the image, gradually guiding the model to reason through and uncover the geographical information. Each set of multi-turn QAs includes the following key attributes: \textit{question ID, source dataset, image path, three geo-questions w/ reasoning process, and one loc-question w/ ultimate answer}. This structure ensures the logical coherence of the multi-turn QAs and clearly presents the progression from question to reasoning process to the final answer.

Prompting techniques improve LLMs' reasoning and problem solving abilities across diverse tasks ~\cite{zero-shot-planners, Swe-bench, Vipergpt}.
We integrate the images’ unified metadata format to generate high-quality dialog data. Using the Interactive Reasoning CoT method, we activate two tasks: \textit{CoT Deduction} and \textit{CoT Introspection}. In the next part, we elaborate on the implementation details of these two tasks.

\textbf{CoT Deduction.} To extract the reasoning chain behind the geographic location predictions from GPT-4V as the training data, we explicitly extract the reasoning chain supporting the model’s QA process. Specifically, we draw on the concept of interactive reasoning from reinforcement learning and propose the \textit{CoT Deduction} method to handle the geographic location prediction task. 

The \textit{CoT Deduction} consists of three parts: \textit{(1) Role Setting}. In CoT Deduction, we set up two roles: \textit{Geo-Guessr player} and \textit{questioner}. The questioner and player interact, with the questioner asking questions and the player responding based on the image clues and existing knowledge. 
\textit{(2) Reasoning QA}. We aim to explicitly extract the internal principles of geographic location reasoning to construct MG-Geo's Dialog Part. For each question from the questioner, the Geo-Guessr player gradually deduces the geographic location based on various aspects embedded in the image, such as the environment and climate, architecture and landmarks, language and culture, and people’s appearance. Each QA round (\textit{i.e.}, Q1A1, Q2A2, and Q3A3) helps the player narrow down the possibilities, gradually approaching the correct answer.
\textit{(3) Coordinate Prediction}. After a series of reasoning steps, the player needs to provide a specific geographic coordinate and briefly explain their choice (Q4A4). 

After \textit{CoT Deduction} generates the predicted coordinates, we initiate a \textit{Decision Criterion} to evaluate the predicted coordinates’ accuracy. Specifically, we calculate the Haversine distance between the predicted coordinates and the unified metadata. If the distance between the predicted and true coordinates is greater than $25km$, the \textit{CoT Introspection} process is triggered.

\textbf{CoT Introspection.}
If the initial coordinate prediction deviates from the metadata by more than 25km, we trigger CoT Introspection. Instead of treating the ground-truth location as a shortcut answer, this stage asks the MLLM to re-check whether each reasoning step is visually grounded, whether the selected clues are geographically discriminative, and whether any high-specificity evidence has been overlooked. The resulting sample is retained only when the revised reasoning remains supported by visible image evidence and passes the quality-control criteria. We explicitly tag all such samples as \texttt{introspection}; in MG-Geo, 70.1\% of the Dialog Part, approximately 51K/73K samples, follows this path. On the triggered subset, the original pre-refinement deduction already reaches 85.2\% Acc@750km and 60.2\% Acc@200km, suggesting that Introspection mainly refines directionally correct but insufficiently fine-grained reasoning rather than replacing arbitrary predictions.

\section{The GaGA Model}
\begin{figure}[h] 
    \centering
    \includegraphics[width=1\linewidth]{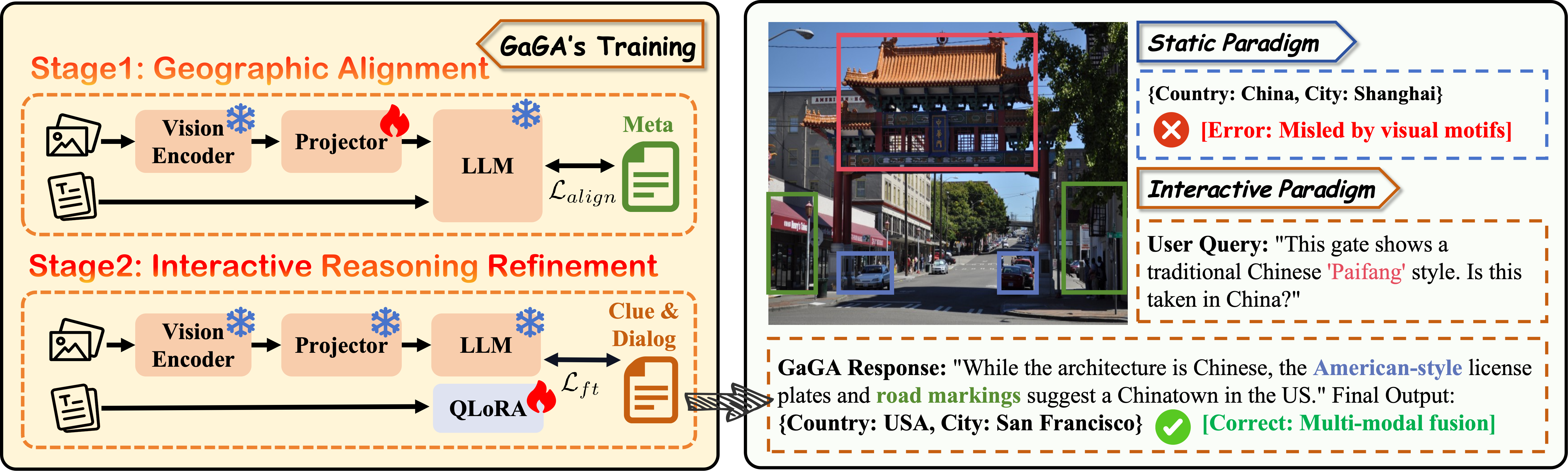} 
    \caption{\textbf{Overview of the GaGA framework.} We transition the geolocation task from a static classification paradigm to a dynamic interactive paradigm, enabling the model to refine predictions through logical constraints. The architecture utilizes a dual-stream training process to bridge the gap between visual motifs and geographic knowledge.}
    \label{gaga}
\end{figure}

Capitalizing on the MG-Geo dataset, we present GaGA, an interactive MLLM designed to transcend the \textit{"black box"} nature of traditional geolocation. As illustrated in Figure~\ref{gaga}, GaGA shifts the geolocation paradigm from static coordinate regression to a dynamic reasoning process. By integrating robust spatial awareness with extensive world knowledge, GaGA can effectively leverage logical constraints during user interaction to refine its predictions in real-time.
\subsection{Model Architecture}
GaGA adopts a modular MLLM architecture following the LLaVA paradigm~\cite{llava}, optimized for iterative spatial reasoning. The framework integrates a vision encoder $f_{V}$, a cross-modal projector $f_P$, and an LLM $f_L$ (Llama3-8B~\cite{llama3}) as the central reasoning engine.

\textbf{Visual Encoding and Projection.} 
For an input street-view image $X_v$, we employ a pretrained CLIP-ViT~\cite{vit} as the vision encoder $f_{V}$ to extract high-level semantic features. To bridge the modality gap, the projector $f_P$ (a two-layer MLP) maps these visual representations into the LLM's latent word embedding space. The resulting visual tokens are obtained via $E_v = f_{P}(f_{V}(X_v))$. 

\textbf{Multimodal Contextualization and Reasoning.}
To support the interactive geolocation paradigm, the text tokenizer $f_T$ is designed to process not just a single query, but a progressive reasoning trajectory. It transforms the cumulative dialogue history $H_{<n}$ and the current user query $Q_n$ (which may contain high-specificity geographic priors, such as language or traffic rules) into textual tokens $E_{T,n} = f_T([H_{<n}; Q_n])$. These tokens are concatenated with $E_v$ to form a unified input sequence. The LLM then performs auto-regressive inference to generate a refined reasoning trajectory $R_n = f_{L}([E_v; E_{T,n}])$. 
This architecture facilitates a deep cross-modal exchange, allowing $f_L$ to contextualize abstract visual patterns within the provided geographic constraints, sequentially updating its belief state to escape visual ambiguities.

\subsection{Training Strategy and Objectives}
The training protocol of GaGA is conceptualized as a two-stage paradigm: \textit{Geographic Alignment} and \textit{Interactive Instruction Tuning}. This progressive approach ensures that the model first internalizes a global geographic world-view before mastering complex spatial reasoning and dialogue capabilities.

\textbf{Stage 1: Geographic Alignment.} 
The primary objective of this stage is to bridge the modality gap between generic visual features and structured geographic metadata. We initialize the multimodal projector with ShareGPT4V~\cite{sharegpt} weights to leverage its robust cross-modal mapping. During this phase, both the vision encoder and the LLM backbone are kept frozen to preserve their respective pre-trained representational power. 

We optimize the projector $\phi$ by minimizing the standard cross-entropy loss over the \textit{Meta Part} of the MG-Geo dataset. Formally, given an image $X_v$ and its associated geographic metadata $Y_{geo}$ (comprising country, region, city, and coordinates), the training objective is:
\begin{equation}
    \mathcal{L}_{align} = -\sum_{t=1}^{T} \log P(y_t | X_v, y_{<t}; \phi),
\end{equation}
where $Y_{geo} = \{y_1, ..., y_T\}$ represents tokenized geographic attributes. This stage forces the projector to learn a \textit{Spatial-Aware Embedding} that aligns visual patterns with specific global hierarchies.

\textbf{Stage 2: Interactive Reasoning Refinement.} 
The second stage focuses on cultivating \textbf{Constraint-based Reasoning} and multi-turn adaptability. We freeze the projector $\phi$ and apply QLoRA~\cite{QLoRA} to the LLM to optimize its decision-making logic. The training corpus integrates the Meta, Clue, and Dialog parts of MG-Geo, with a strategic emphasis on the \textit{Interactive Reasoning CoT} strategy. Crucially, the training corpus here is a heterogeneous mixture of 240k carefully curated samples from the Meta, Clue, and Dialog parts of MG-Geo. The \textit{Clue Part} explicitly exposes the model to 8 distinct dimensions of geographic evidence (e.g., architecture, vegetation, road markings), while the \textit{Dialog Part} instills the \textit{Interactive Reasoning CoT} strategy. The optimization goal is to maximize the conditional likelihood of the correct reasoning chains $R$ and the final coordinates $C$ given the visual input $X_v$ and user queries $Q$:
\begin{equation}
    \mathcal{L}_{ft} = -\mathbb{E}_{(X_v, Q, R, C) \sim \mathcal{D}_{MG-Geo}} [ \log P(R, C | X_v, Q; \theta_{LoRA}) ].
\end{equation}

By tightly coupling the LLM optimization with MG-Geo's conversational structure, this multi-task objective enables GaGA to identify and rectify its initial misjudgments. This ensures that the final response is the result of a deliberate, multi-turn evidentiary synthesis rather than a simplistic statistical mapping.

\section{Experiment}
To demonstrate the efficacy of our dataset in addressing the geographic knowledge gap in existing models and to showcase GaGA's ability in geolocation tasks, we conducted a comprehensive suite of experiments, the primary findings of which are presented in this section.

\subsection{Experimental Setup}\label{exp setting}

\textbf{Metrics.}
We use the following metrics to evaluate the prediction accuracy of the geolocation model:

(1) Accuracy of predicted geographical names at various administrative levels: country, region and city.

(2) Accuracy of predicted coordinates within various distance thresholds: 1km, 25km, 200km, 750km, and 2500km, calculated as the haversine distance between the model's predicted GPS coordinates and the ground truth.

(3) Geoscore: it is defined as  $5000exp(-\delta/1492.7)$ based on the famous ``Geoguessr'' game. $\delta$ represents the Haversine distance between predicted and ground truth image locations.

\textbf{Benchmark.} We assess GaGA's performance across multiple benchmarks. For its primary global geolocation capability, we use GWS15k~\cite{gws15k} and OSV-5M-test~\cite{OSV5M} for its balanced worldwide coverage. Since GWS15k is not open-source, we reproduce it in this study. The pseudocode and data distribution are presented in the Appendix~\ref{app:gws15k_reproduction}. To further evaluate its generalization capability, we conduct comparative experiments between GaGA and state-of-the-art methods on two web-crawled image benchmarks, Im2GPS3k~\cite{vo2017revisiting} and YFCC4k~\cite{yfcc100m}, with results reported in the Appendix~\ref{app:standard_benchmarks}.

\textbf{Implementation Details.} All experiments are conducted using the XTuner platform~\cite{xtuner}, facilitating efficient multimodal model tuning and deployment. For reasoning tasks, we use LMDeploy~\cite{lmdeploy} tool. We conduct all the experiments on 8 $\times$ RTX4090 GPUs.

\subsection{Geolocation Performance}
\label{Geolocation Performance}
The evaluation results on the GWS15k benchmark are summarized in Table~\ref{tab: Geolocation performace}, where we categorize current methodologies into traditional discriminative models and MLLM-based reasoning assistants.
\begin{table}[!t]
    \centering
    \caption{\textbf{Administrative-Level Accuracy and Coordinates Accuracy of GaGA on GWS15K Bench. Left:} Administrative-Level. \textbf{Right:} Coordinates Accuracy.  We use \textbf{bold} to indicate the best performance, and `\underline{\phantom{XX}}' for the second-best, respectively. $^\bigstar$ represents the model evaluated on GWS15k reproduced in this paper.}
    \begin{minipage}[b]{0.43\textwidth}
        \centering
        \setlength\tabcolsep{3pt}
        \renewcommand{\arraystretch}{1.2}
        \resizebox{\linewidth}{!}{
            \begin{tabular}{l|c|cc}
            \toprule
            \multirow{2}{*}{\textbf{Method}} & \multirow{2}{*}{\textbf{Recall}} & \multicolumn{2}{c}{\textbf{Admin-Level(\%)}} \\
            & & Country & City \\
            \specialrule{\lightrulewidth}{0pt}{0pt}
            StreetCLIP\cite{streetclip} & 1 & 40.11 & 3.02 \\
            Hybrid\cite{OSV5M} & 1 & \textbf{58.49} & \textbf{3.36} \\
            \midrule
            LLaVA-Llama3 & 0.99 & 1.76 & 0.02 \\
            InternVL2-8B\cite{Internvl2} & 0.96 & 24.74 & 0.48 \\
            Qwen2.5-VL-7B\cite{Qwen2.5-VL} & 0.99 & 33.37 & 1.43 \\
            Qwen3-VL-8B\cite{Qwen3-VL} & 0.99 & 39.10 & 1.63 \\
            GeoReasoner\cite{GeoReasoner} & 1 & 40.63 & 1.11 \\
            GLOBE\cite{GLOBE} & 1 & \underline{47.09} & \underline{3.44} \\
            GaGA(Ours) & 1 & \textbf{63.06} & \textbf{6.28} \\
            \specialrule{\heavyrulewidth}{0pt}{0pt}
        \end{tabular}
        }
    \end{minipage}
    \hfill
    \begin{minipage}[b]{0.55\textwidth}
        \centering
        \renewcommand{\arraystretch}{1.2} 
        \resizebox{\linewidth}{!}{
            \begin{tabular}{l|ccccc|c}
                \toprule
                \multirow{2}{*}{\textbf{Method}} &  \multicolumn{5}{c|}{\textbf{Coordinates Accuracy  (\% @ km)}} &
                \multirow{2}{*}{\textbf{Geoscore}} \\
                ~ & 1km & 25km & 200km & 750km & 2500km & \\ \midrule
                ISNs\cite{ISNs} & 0.05 & 0.6 & 4.2 & 15.5 & 38.5 & - \\
                Translocator\cite{Translocator} & \underline{0.5} & 1.1 & 8 & 25.5 & 48.3 & -\\
                GeoDecoder\cite{GeoDecoder} & \textbf{0.7} & 1.5 & 8.7 & 26.9 & 50.5 & - \\
                GeoCLIP$^\bigstar$\cite{GeoCLIP} & 0.2 & 3.1 & 15.4 & 40.3 & 71.2 & \underline{2345.2}\\
                PIGEOTTO\cite{Pigeon} & \textbf{0.7} & \underline{9.2} & \underline{31.2} & \textbf{65.7} & \textbf{85.1} & -\\
                Hybrid$^\bigstar$\cite{OSV5M} & 0.08 & \textbf{14.9} & \textbf{39.3} & \underline{56.2} & \underline{74.4} & \textbf{2944.9}\\
                \specialrule{\lightrulewidth}{0pt}{0pt}
                \rowcolor{backcolor}
                GeoReasoner$^\bigstar$\cite{GeoReasoner} & 0.04 & 2.1 & 9.8 & 32.9 & 64.0 & 2018.4\\
                \rowcolor{backcolor}
                GRE\cite{gre} & \textbf{0.9} & \underline{4.1} & \underline{18.9} & \underline{54.8} & \underline{78.3} & - \\
                \rowcolor{backcolor}
                GLOBE$^\bigstar$\cite{GLOBE} & \underline{0.1} & 3.9 & 15.0 & 40.8 & 69.4 & \underline{2311.3} \\
                \rowcolor{backcolor}GaGA$^\bigstar$(Ours) & \underline{0.1} & \textbf{8.5} & \textbf{33.9} & \textbf{60.6} & \textbf{82.2} & \textbf{3113.0}\\
                \specialrule{\heavyrulewidth}{0pt}{0pt}
            \end{tabular}
        }
    \end{minipage}
    \label{tab: Geolocation performace}
\end{table}

As shown in Table~\ref{tab: Geolocation performace} (Left), GaGA achieves outstanding performance in administrative boundary identification. Notably, while general-purpose MLLMs (e.g., LLaVA-Llama3, Qwen-VL~\cite{Qwen2.5-VL, Qwen3-VL}) and even specialized geographic MLLMs (e.g., GeoReasoner~\cite{GeoReasoner}, GLOBE~\cite{GLOBE}) struggle with precise localization, GaGA achieves SOTA performance among all MLLM-based methods. Specifically, GaGA attains a country-level accuracy of 63.06\% and a city-level accuracy of 6.28\%, outperforming the previous best-performing MLLM, GLOBE, by a substantial margin of 15.97\% and 2.84\% respectively. Even when compared to traditional SOTA classifiers like Hybrid~\cite{OSV5M}, GaGA maintains a clear lead (+4.57\% at country level and +2.92\% at city level), demonstrating that the interactive reasoning paradigm does not sacrifice, but rather enhances, fundamental localization precision. Furthermore, GaGA maintains a 100\% Recall rate, ensuring that every query is met with a valid geographic coordinate—a critical reliability metric where other LLMs often falter.

Table~\ref{tab: Geolocation performace} (Right) presents the coordinate-level error distribution. Among MLLM-based approaches (highlighted in the bottom section), GaGA consistently defines the state-of-the-art, significantly surpassing GeoReasoner, GRE, and GLOBE across thresholds of 25 to 2500km. For instance, at the 200km threshold—a key metric for regional localization—GaGA leads the second-best MLLM (GRE) by 15.0\%.
When compared to traditional end-to-end models, GaGA exhibits competitive performance, securing the second-best results from 200km to 2500km. Most importantly, GaGA achieves the \textbf{highest Geoscore (3113.0)} among all reproducible models, including the strong Hybrid baseline. This peak Geoscore underscores GaGA’s unique ability to strike an optimal balance between high-precision hits and the mitigation of catastrophic outliers, a direct result of its clue-based reasoning and interactive refinement capabilities.

Table~\ref{tab:comparison_on_osv5m} reports the comparative results on the OSV-5M test set. Notably, GaGA achieves the highest City-level accuracy (7.4\%) among all competing methods. In terms of global metrics, GaGA reaches a Geoscore of 3413 and a Country-level accuracy of 71.4\%, consistently outperforming specialized models like ISNs and the Hybrid baseline. These findings underscore GaGA’s robustness in balancing macro-scale region recognition and micro-scale city identification.

\definecolor{darkgreen}{RGB}{255, 128, 0}
\definecolor{s_blue}{RGB}{0, 153, 0}

\subsection{Inference-Time Interactive Geolocation Analysis}
A fundamental strength of GaGA lies in its capacity for \textbf{Multi-tier Interactive Reasoning}, allowing it to dynamically refine geolocation beliefs through dialogue. In this section, we provide a formal dissection of this capability. First, we establish a taxonomic framework for interactive guidance to quantify performance gains across different levels of information density. Second, we investigate the "Similarity Trap" to elucidate how the specificity of geographic cues influences the model's posterior distribution.

\subsubsection{Effectiveness of Tiered Interactive Guidance}\label{Performance of location in interactive scenarios}
To evaluate GaGA’s reasoning flexibility, we curated a diagnostic set of 547 images spanning diverse cultural and natural landscapes. We standardized the interaction into three progressive tiers: 
(1) \textbf{Zero-shot (Direct Inquiry)}: Baseline prediction without external cues.
(2) \textbf{Tier-1 (Focus-based Inquiry)}: Providing a guiding question (e.g., \textit{``Considering the architectural design, what region of the world would you think displays such forms?''}) to direct the model's attention to specific attributes without providing factual answers.
(3) \textbf{Tier-2 (Knowledge-augmented Guidance)}: Providing both a question and a factual geographic prior (e.g., \textit{``The photo shows the typical oceanic and continental climate zones; which countries typically use this color?''}).

\begin{table}[!t]
    \centering
    \begin{minipage}[t]{0.48\linewidth}
        \centering
        \caption{Performance of GaGA with Tiered Interactive Guidance.}
        \label{tab:geographic knowledge question answering}
        \resizebox{\linewidth}{!}{%
        \begin{tabular}{lccc}
        \toprule
        \multirow{2}{*}{\textbf{Interaction Tiers}} & \multicolumn{3}{c}{\textbf{Admin-Level Accuracy (\%)}} \\
        \cmidrule(lr){2-4}
         & Country & Region & City \\
        \midrule
        Direct Inquiry & 64.89 & 27.97 & 7.67 \\
        \multirow{2}{*}{Focus-based Inquiry} 
        & 61.24 & 29.25 & 8.22\\
        & \textcolor{s_blue}{$-3.65$} & \textcolor{darkgreen}{$+1.28$} & \textcolor{darkgreen}{$+0.55$} \\ 
        \multirow{2}{*}{\makecell[c]{Knowledge-augmented \\ Guidance}} 
        & \textbf{74.77} & \textbf{34.73} & \textbf{9.87}\\
        & \textcolor{darkgreen}{$+9.88$} & \textcolor{darkgreen}{$+6.76$} & \textcolor{darkgreen}{$+2.20$} \\
        \bottomrule
        \end{tabular}%
        }%
    \end{minipage}%
    \hfill
    \begin{minipage}[t]{0.43\linewidth}
        \centering
        \caption{Impact of Clue Ambiguity on Geolocation Performance.}
        \label{tab:ambiguity_impact}
        \renewcommand{\arraystretch}{1.2} 
        \resizebox{\linewidth}{!}{%
        \begin{tabular}{lccc}
            \toprule
            \multirow{2}{*}{\textbf{Type of Subset}} & \multicolumn{3}{c}{\textbf{Admin-Level Accuracy (\%)}} \\
            \cmidrule(lr){2-4}
             & Country & Region & City \\
            \midrule
            Ambiguous & 59.5 & 26.2 & 7.0 \\
            High-Specificity & \textbf{62.2} & \textbf{28.9} & \textbf{9.1} \\
            \bottomrule
        \end{tabular}%
        }%
    \end{minipage}
\end{table}

Table \ref{tab:geographic knowledge question answering} illustrates the performance dynamics. The results reveal several key insights. 
Firstly, \textbf{Tier-2 interaction} yields the most significant gains, with GaGA's country-level accuracy surging by 9.88\% (reaching 74.77\%). This demonstrates GaGA's superior ability in \textit{Knowledge Integration}—effectively using external anchors to prune the search space. 
Secondly, an intriguing observation arises in \textbf{Tier-1 interaction}: while region and city-level precisions improve, country-level accuracy experiences a slight decline (-3.65\%). We characterize this as the exploration cost within the \textit{Similarity Trap}. Without a factual anchor, Tier-1 queries compel the model to focus on broadly distributed visual features. While this sharpens the model's focus on local details (improving city-level hits), it simultaneously increases the risk of being misled by cross-border visual similarities. 

\begin{figure}[!t] 
\centering
\includegraphics[width=1\linewidth]{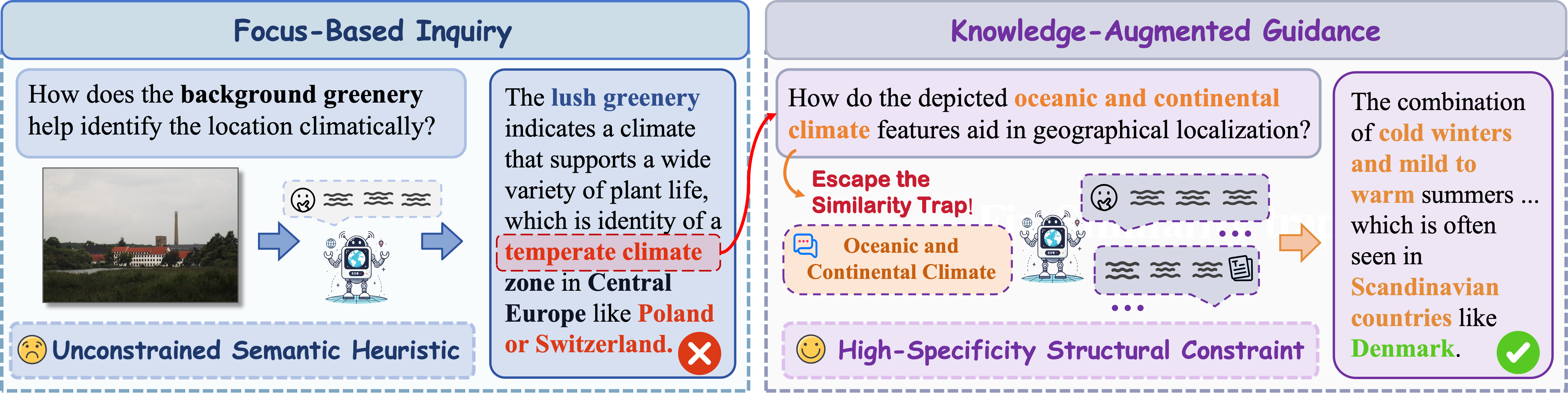}
\caption{Navigating through the ``Similarity Trap'' via tiered interaction. }
\label{fig:interaction}
\end{figure}

\subsubsection{Geographic Ambiguity and the ``Similarity Trap''}\label{sec:case_study_ambiguity} 
Global geolocation is inherently plagued by visual ambiguity, a challenge we formalize as the \textbf{``Similarity Trap''}. This occurs when geographically widespread features—such as neoclassical architecture or generic temperate flora—produce overlapping representations in the model's feature space, leading to over-confident misclassifications~\cite{around_the_world_in_80_timesteps}. While static retrieval-based models are often paralyzed by such ambiguity, GaGA’s interactive framework introduces a dynamic \textit{Error-Correction Mechanism}.

As illustrated in the comparative analysis in Figure~\ref{fig:interaction}, when GaGA is prompted with an \textbf{Unconstrained Semantic Heuristic} (e.g., \textit{``How does the background greenery help identify the location?''}), it may initially succumb to the trap by over-relying on generic motifs that lack geographic exclusivity. However, by transitioning to a \textbf{High-Specificity Structural Constraint} (e.g., \textit{``How do the oceanic and continental climate features aid in localization?''}), the model is forced to decouple ambiguous visual signals and engage in constraint-based reasoning. This logical anchoring allows GaGA to suppress noise and re-route its reasoning toward the correct manifold.

Formally, let $G$ denote the geographic label and $c$ a visual clue. We define clue specificity as
\[
s(c) = 1 - \frac{H(G|c)}{H(G)},
\]
where low-specificity clues have high conditional entropy over possible locations, while high-specificity clues sharply reduce the uncertainty. Given model salience weights $\alpha_i$ over visual clues $\{c_i\}$, a Similarity Trap occurs when the prediction is dominated by low-specificity clues, i.e., $\sum_i \alpha_i(1-s(c_i))$ is high, leading to a localization error $d(\hat{y}, y^*)>\tau$.

Based on this specificity criterion, we partitioned our evaluation set into two diagnostic subsets based on the information entropy and geographic exclusivity of the cues:

\textbf{(1) High-Specificity Subset:} Focuses on \textit{Deterministic Markers} with narrow geographic distributions and strong logical constraints (\textit{e.g.}, road markings, signage language, or driving side).

\textbf{(2) Ambiguous Subset:} Focuses on \textit{Distributive Features} with high spatial variance and weak exclusionary power (\textit{e.g.}, architectural styles, general vegetation, or terrain).

The empirical results in Table~\ref{tab:ambiguity_impact} confirm that high-specificity cues yield significantly higher precision across all administrative levels. This validates our hypothesis that the ``Similarity Trap'' is the primary bottleneck in geolocation. More importantly, it underscores the necessity of GaGA’s interactive paradigm: by converting human-provided high-specificity priors into logical constraints, GaGA effectively navigates through visual ambiguity where static models fail.

\subsection{Ablation Experiments}\label{ablation study}

To systematically evaluate the contribution of each component within the MG-Geo dataset, we conduct a comprehensive ablation study. Using the test set from Section~\ref{Performance of location in interactive scenarios}, we design five experimental groups. We evaluate performance across four key metrics: \textit{Accuracy} (Geo-Score), \textit{Informativeness} (hit rate of generated clues, assessed by an LLM Judge), \textit{Fluency} (Likert scale score for coherence, assessed by an LLM Judge), and \textit{Relevance} (contextual relevance, assessed by an MLLM Judge). 

As detailed in Table~\ref{tab:ablation}, our ablation study validates the indispensable roles and intentional design trade-offs of the MG-Geo dataset components. The \texttt{Meta} data serves as the bedrock for static localization, achieving the highest overall Accuracy (3872). Conversely, the \texttt{Dialog} component acts as the primary driver for conversational reasoning, yielding substantial improvements across interactive metrics, including a significant leap in Informativeness (from 0.6164 to 0.7549). Furthermore, the \texttt{Clue} data functions as a conditional enhancer; while it can slightly perturb accuracy when lacking conversational context, combining it with the reasoning framework (\texttt{Meta}+\texttt{Clue}+\texttt{Dialog}) maximizes both Informativeness (0.7809) and Fluency (4.9555). Ultimately, this deliberate trade-off—sacrificing a marginal degree of raw precision for massive gains in explainability and interaction—confirms our core objective of developing a balanced, interactive global geolocation assistant.

We further isolate the effect of CoT Introspection by removing all Introspection-tagged dialogs while maintaining the same Stage-2 sample budget. The full model improves over the w/o-Introspection variant by +5.74 Acc@200km, +12.77 Acc@750km, +15.00 Acc@2500km, and +7.23 country-level accuracy, supporting the role of Introspection as a cue-correction signal rather than a coordinate shortcut. Detailed results are provided in the Appendix~\ref{app:introspection}.

\begin{table}[!t]
    \centering
    \begin{minipage}[t]{0.52\linewidth}
        \centering
        \caption{Ablation study on the components of the MG-Geo dataset.}
        \label{tab:ablation}
        \resizebox{\linewidth}{!}{
        \begin{tabular}{lcccc}
            \toprule
            \textbf{Training Data} & \textbf{Acc.} & \textbf{Info.} & \textbf{Flu.} & \textbf{Rel.} \\
            \midrule
            Meta & \textbf{3872} & 0.616 & 4.572 & 2.294 \\
            Meta+Clue & 3835 & 0.617 & 4.494 & 2.168 \\
            Meta+Dialog & 3854 & 0.754 & 4.929 & 3.579 \\
            Meta+Clue+Dialog & 3817 & \textbf{0.780} & \textbf{4.955} & 3.650 \\
            Clue+Dialog & 3773 & 0.767 & 4.925 & \textbf{3.898} \\
            \bottomrule
        \end{tabular}
        }
    \end{minipage}
    \hfill 
    \begin{minipage}[t]{0.4\linewidth}
        \centering
        \caption{Performance comparison on OSV-5M-test set.}
        \label{tab:comparison_on_osv5m}
        \resizebox{\linewidth}{!}{
        \begin{tabular}{l|c|cc}
            \toprule
            \multirow{2}{*}{\textbf{Model}} & \multirow{2}{*}{\textbf{Geoscore}} & \multicolumn{2}{c}{\textbf{Admin-Level}} \\
             &  & Country & City \\ 
            \midrule
            ISNs\cite{ISNs} & 3331 & 66.8 & 4.2 \\
            Hybrid\cite{OSV5M} & 3361 & 67.4 & \underline{6.0} \\
            RFM $\mathcal{S}_2$\cite{around_the_world_in_80_timesteps} & \textbf{3767} & \textbf{76.2} & 5.4 \\
            GaGA(Ours) & \underline{3413} & \underline{71.4} & \textbf{7.4} \\ 
            \bottomrule
        \end{tabular}
        }
    \end{minipage}
\end{table}

\subsection{Release, Privacy, and Responsible Use}
MG-Geo is released in a source-compatible form, including annotations, prompts, splits, generation-path tags, and reconstruction scripts.

\section{Conclusion}
In this work, we address key challenges in global geolocation by tackling the critical lack of comprehensive data and the performance limitations of existing methods. To this end, we introduce \textbf{MG-Geo}, the first large-scale, high-quality multimodal dataset specifically designed with rich geographic cues to bridge this knowledge gap for MLLMs. Building upon this foundational dataset, we developed \textbf{GaGA}, a novel MLLM that not only demonstrates superior performance over existing models in predicting administrative boundaries but also introduces a crucial interactive capability for refining localization through user dialogue. Our work underscores the critical role of high-quality, domain-specific datasets in unlocking the potential of MLLMs for complex spatial reasoning tasks and opens new avenues for a wide range of geographic downstream applications.

\section*{Acknowledgements}
This work was supported in part by the Key Deployment Program of the Chinese Academy of Sciences, China under Grant KGFZD-145-25-39, the National Natural Science Foundation of China under Grants 62272438, and Beijing Natural Science Foundation L25700.

\newpage
\bibliographystyle{splncs04}
\bibliography{mybibfile}

\clearpage
\appendix
\section*{Appendix}
\section{CoT Introspection Analysis}
\label{app:introspection}

\paragraph{Trigger and tagging.}
After CoT Deduction predicts coordinates, we compute the Haversine distance between the prediction and the metadata coordinates. CoT Introspection is triggered when this distance exceeds $25\,\mathrm{km}$. In MG-Geo, 70.1\% of the Dialog Part, approximately 51K out of 73K samples, follows this path. We explicitly tag every Dialog sample with its generation path, either \texttt{deduction} or \texttt{introspection}, so downstream users can include, exclude, or separately analyze Introspection-generated samples.

\paragraph{Role of Introspection.}
Simply feeding a ground-truth answer back to an MLLM can lead to post-hoc rationalization if left uncontrolled. Our intended use of Introspection is therefore evidence re-checking rather than answer-forced explanation. Given a verified geographic anchor, the model is asked to re-examine whether each reasoning step is visually grounded, whether the selected clues are geographically discriminative, and whether high-specificity evidence has been overlooked. 

On the triggered subset, the original pre-refinement deduction already reaches 85.2\% Acc@750km and 60.2\% Acc@200km. This indicates that Introspection mainly refines directionally plausible but insufficiently fine-grained reasoning, rather than replacing arbitrary predictions with a known answer. No ground-truth coordinates are provided during downstream model evaluation.

\begin{table}[ht]
  \centering
  \caption{Impact of CoT Introspection under a size-matched Stage-2 training budget.}
  \label{tab:ablation_introspection}
  \small
  \resizebox{\linewidth}{!}{%
  \begin{tabular}{lccccccc}
    \toprule
    \textbf{Stage-2 Data} & \textbf{Acc@25} & \textbf{Acc@200} & \textbf{Acc@750} & \textbf{Acc@2500} & \textbf{Country} & \textbf{Region} & \textbf{City} \\
    \midrule
    Full MG-Geo & 17.96 & 41.30 & 78.70 & 93.89 & 61.30 & 27.78 & 8.15 \\
    w/o Introspection & 17.41 & 35.56 & 65.93 & 78.89 & 54.07 & 22.78 & 7.96 \\
    $\Delta$ & +0.55 & +5.74 & +12.77 & +15.00 & +7.23 & +5.00 & +0.19 \\
    \bottomrule
  \end{tabular}}
\end{table}

The ablation removes all Introspection-tagged dialogs while maintaining the same Stage-2 sample budget by resampling non-Introspection Clue/metadata samples under the identical training recipe. The full model improves over the w/o-Introspection variant most strongly at coarse-to-mid coordinate thresholds, suggesting that Introspection functions as a cue-correction signal that reduces catastrophic far-away predictions.

\section{Geographic Distribution and Regional Performance}
\label{app:geo_distribution}

\begin{figure*}[t]
    \centering
    \includegraphics[width=0.95\linewidth]{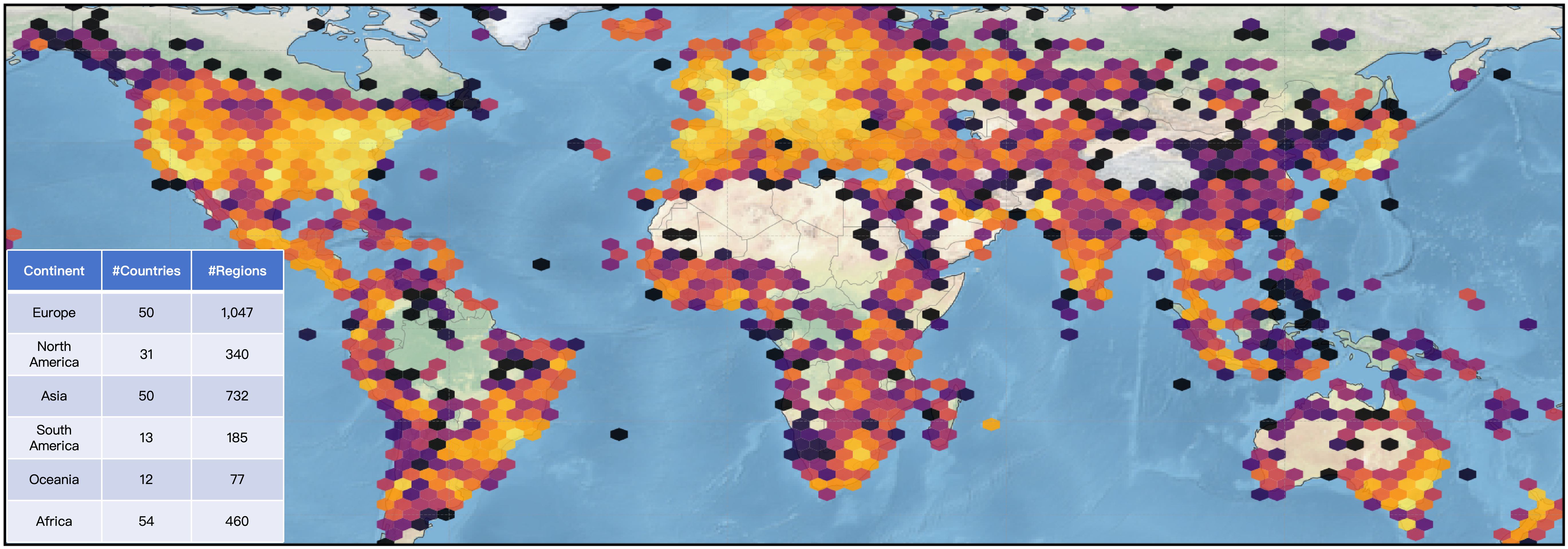}
    \caption{Global coordinate distribution of MG-Geo. The dataset has broad coverage across countries and territories but remains naturally long-tailed due to uneven street-view and landmark availability.}
    \label{fig:data_heatmap}
\end{figure*}

\begin{figure*}[t]
    \centering
    \includegraphics[width=0.95\linewidth]{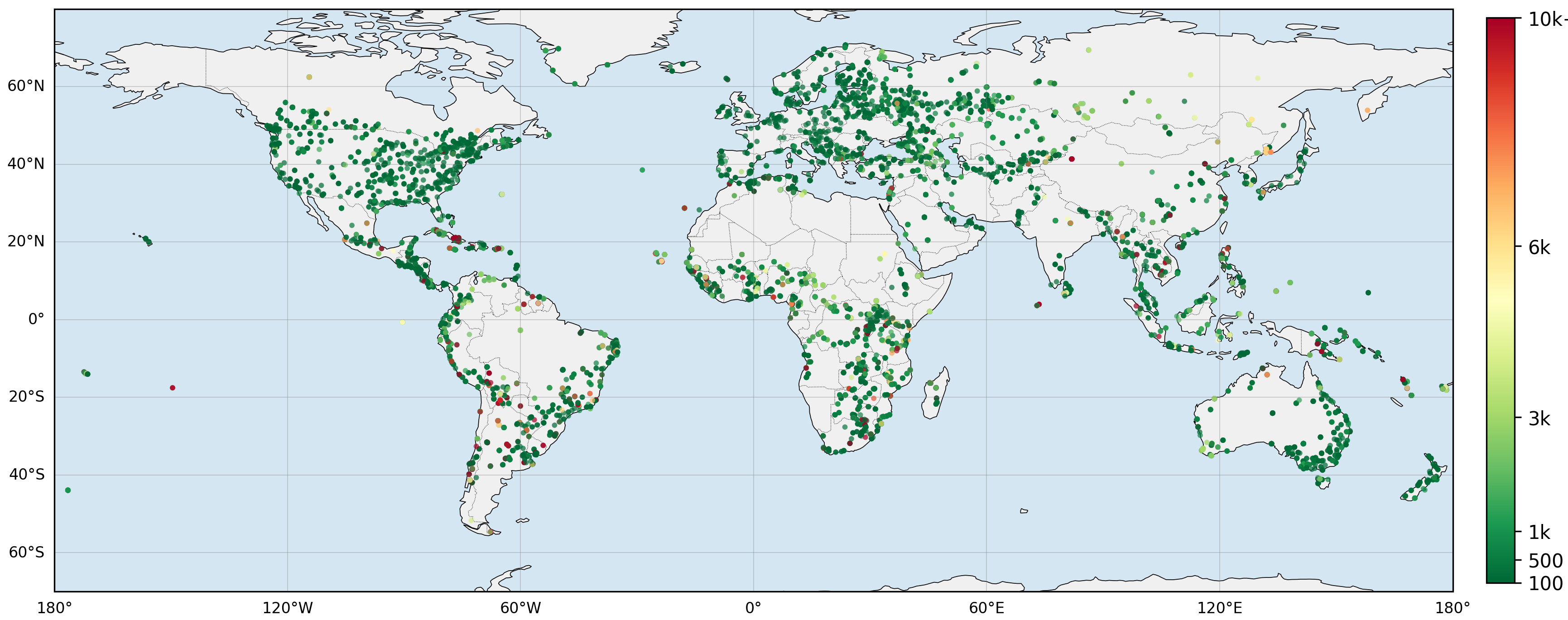}
    \caption{Geodesic error distribution of GaGA on the reproduced GWS15k-style benchmark. Regions with sparse data and visually homogeneous scenes remain more challenging, motivating the use of high-specificity anchors in interactive evaluation.}
    \label{fig:error_heatmap}
\end{figure*}

MG-Geo provides broad geographic coverage but naturally follows a long-tailed distribution inherited from global street-view and landmark availability. To make this distribution transparent, we report both sample density and regional error patterns. Fig.~\ref{fig:data_heatmap} visualizes the global coordinate distribution of MG-Geo, while Fig.~\ref{fig:error_heatmap} shows the geodesic error distribution of GaGA on the reproduced GWS15k-style benchmark.

We release the full per-country distribution as a CSV file together with the dataset card. MG-Geo is characterized as a broad-coverage but long-tailed resource, and this distribution is exposed for downstream analysis.

\section{Backbone Transferability}
\label{app:backbone_transfer}

To verify that the gains of MG-Geo are not tied to a single LLaVA-Llama3 backbone, we fine-tune Qwen3-VL-8B with the same MG-Geo supervision recipe and evaluate it on the same reproduced GWS15k-style benchmark. Table~\ref{tab:transfer_qwen} shows that MG-Geo supervision provides large gains across both coordinate-level and administrative-level metrics.

\begin{table}[t]
  \centering
  \caption{Transfer learning results on Qwen3-VL-8B.}
  \label{tab:transfer_qwen}
  \small
  \resizebox{\linewidth}{!}{%
  \begin{tabular}{lccccccc}
    \toprule
    \textbf{Model / Config} & \textbf{Acc@25} & \textbf{Acc@200} & \textbf{Acc@750} & \textbf{Acc@2500} & \textbf{Country} & \textbf{Region} & \textbf{City} \\
    \midrule
    Qwen3-VL-8B & 1.16 & 8.10 & 28.75 & 59.95 & 39.10 & 5.86 & 1.63 \\
    + MG-Geo SFT & 12.78 & 37.41 & 61.38 & 81.41 & 66.74 & 29.69 & 7.46 \\
    $\Delta$ & +11.62 & +29.31 & +32.63 & +21.46 & +27.64 & +23.83 & +5.83 \\
    \bottomrule
  \end{tabular}}
\end{table}

For Qwen3-VL-8B, Stage 1 trains only the multimodal projector on the Meta Part. Stage 2 starts from the Stage-1 checkpoint, freezes the vision encoder and multimodal projector, and applies LoRA to the language model using the Dialog Part. This mirrors the two-stage MG-Geo training protocol and isolates the contribution of the dataset from the original GaGA backbone.

\section{Release Plan, License Compatibility, and Responsible Use}
\label{app:release}

\paragraph{Release scope.}
We release MG-Geo in a source-compatible form, including annotations, train/validation/test splits, prompts, generation-path tags, and reconstruction scripts. Raw images remain governed by the licenses and access terms of their original sources.

\paragraph{Generation-path transparency.}
All Dialog samples are tagged with their generation path, either \texttt{deduction} or \texttt{introspection}. This allows downstream users to filter the dataset according to their tolerance for Introspection-generated reasoning chains.

\paragraph{License compatibility.}
MG-Geo combines geo-tagged street-view metadata and landmark imagery from existing sources. We do not re-license raw images. Instead, we provide source identifiers and reconstruction scripts where appropriate. AI-generated annotations are explicitly marked and released subject to applicable terms of the data sources and model providers.

\paragraph{Privacy and dual-use.}
Street-view imagery may contain sensitive visual content such as faces, license plates, private property, or location traces. MG-Geo is intended for research on geolocation reasoning, benchmark evaluation, and spatial understanding. It should not be used for identifying individuals, tracking private persons, targeted surveillance, or other privacy-invasive applications. We recommend downstream users follow the privacy handling and removal mechanisms of the original image sources.

\section{Reproduction of the GWS15k-style Benchmark}
\label{app:gws15k_reproduction}

Because the original GWS15k split is not publicly released, we construct a GWS15k-style benchmark from OSV-5M-Test for protocol-level comparison. Results on this reproduced split are marked with $^\bigstar$ in the main paper and should not be interpreted as exact split-level comparisons to prior GWS15k reports.

\begin{figure}[h]
    \centering
    \includegraphics[width=0.8\linewidth]{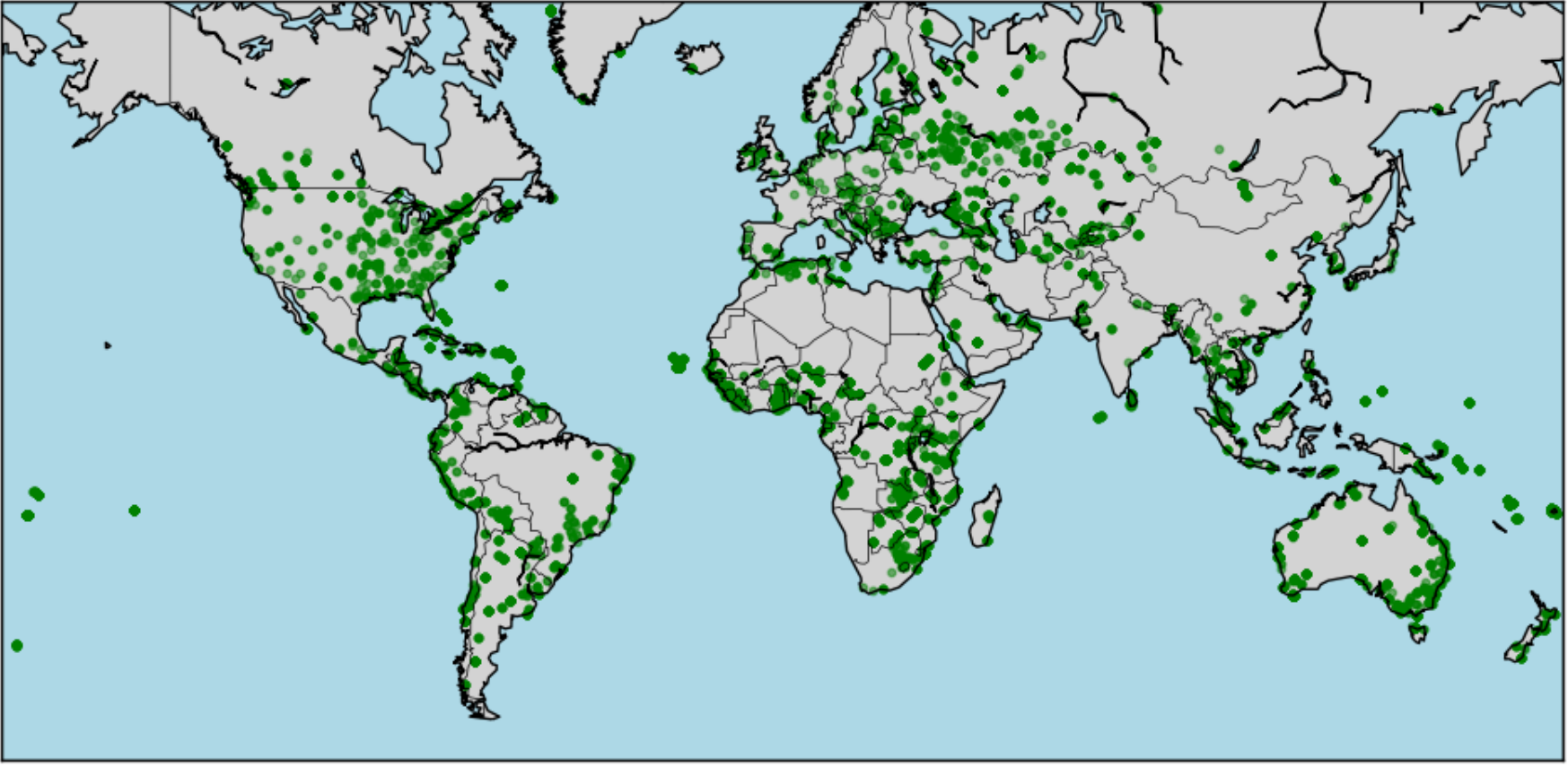}
    \caption{Distribution of the GWS15k-style benchmark reproduced in this work.}
    \label{fig:gws15k_dist}
\end{figure}

We use the test set of OSV-5M as the database and collect geographically distributed imagery based on 43K cities and the surface area of each country. We first sample countries/regions based on their proportion of Earth's surface area, then randomly select a city within each and GPS coordinates within a 5 km radius of that city's center to sample from OSV-5M-Test. The pseudocode is presented in Algorithm~\ref{alg:gen_valid_locs}, and the sampled data distribution is illustrated in Fig.~\ref{fig:gws15k_dist}.

\begin{algorithm}[h]
\caption{Reproduction of the GWS15k-style benchmark}
\label{alg:gen_valid_locs}
\KwIn{Cities dataset $C$, countries dataset $Co$, GPS coordinates $Coord$; maximum number of valid locations $N_{\max}$; radius $R = 5\,\mathrm{km}$.}
\KwOut{Set of valid locations $V$.}

\BlankLine
\textbf{Step 1: Adjusted country probabilities.} \\
\ForEach{country $c \in Co$}{
    $P_{\text{base}}[c] \gets \dfrac{\text{Area}[c]}{A_{\text{total}}}$\;
    $P_{\text{adj}}[c] \gets 0.5 \cdot P_{\text{base}}[c] + \dfrac{0.5}{|Co|}$\;
}

\BlankLine
\textbf{Step 2: Sampling of valid locations.} \\
Initialize $V \gets \emptyset$\;
\While{$|V| < N_{\max}$}{
    Normalize $P_{\text{adj}}$ to sum to 1\;
    Sample a country $c_s \sim P_{\text{adj}}$\;
    Let $S \gets \{\text{city} \in C \mid \text{city.country} = c_s\}$\;
    Sample a city $s_s \in S$\;
    Let $coord_c \gets s_s.\text{coordinates}$\;
    \ForEach{$coord \in Coord$}{
        Compute $d \gets \text{haversine}(coord_c, coord)$\;
        \If{$d \leq R$ \textbf{and} $coord \notin V$}{
            Add $coord$ to $V$\;
        }
    }
}
\Return{$V$}\;
\end{algorithm}

\section{Additional Standard Benchmarks}
\label{app:standard_benchmarks}

When evaluating generalization on Im2GPS3k and YFCC4k, it is important to acknowledge their data distribution limitations. A substantial proportion of images within these datasets lack discernible geographic cues, so performance metrics can be influenced by dataset-specific biases rather than only by spatial reasoning ability~\cite{OSV5M}.

Because Im2GPS3k and YFCC4k contain many non-street-view and retrieval-dependent cases, we additionally report a retrieval-augmented variant, \textbf{GaGA (MP16-Pro)}, for reference. This version equips GaGA with a Retrieval-Augmented Generation module that uses the large-scale MP16-Pro dataset as an external knowledge base. These results are intended to show compatibility with external retrieval resources rather than to replace the main street-view geolocation evaluation.

The performance comparisons on Im2GPS3k and YFCC4k are presented in Table~\ref{tab:im2gps3k}. We categorize the baselines into \textit{Traditional Methods} and \textit{Generation-based Methods}. GaGA (MP16-Pro) improves substantially over the base GaGA model, especially on retrieval-dependent cases. This suggests that a reasoning-focused model can serve as a strong foundation when combined with external retrieval tools.

\begin{table}[t]
    \centering
    \caption{Coordinate accuracy (\% @ km) of GaGA and geolocation models on Im2GPS3k and YFCC4k.}
    \label{tab:im2gps3k}
    \footnotesize
    \setlength{\tabcolsep}{3pt}
    \resizebox{\textwidth}{!}{%
    \begin{tabular}{lccccccccccc}
    \toprule
    \multicolumn{1}{c}{} & \multicolumn{5}{c}{\textbf{Im2GPS3k}} & & \multicolumn{5}{c}{\textbf{YFCC4k}} \\
    \cmidrule(lr){2-6} \cmidrule(l){8-12}
    \textbf{Model} & 1km & 25km & 200km & 750km & 2500km & & 1km & 25km & 200km & 750km & 2500km \\
    \midrule
    \multicolumn{12}{l}{\textit{Traditional Methods}} \\
    Translocator~\cite{Translocator} & 11.8 & 31.1 & 46.7 & 58.9 & 80.1 & & 8.4 & 18.6 & 27.0 & 41.1 & 60.4 \\
    GeoDecoder~\cite{GeoDecoder} & 12.8 & 33.5 & 45.9 & 61.0 & 76.1 & & 10.3 & 24.4 & 33.9 & 50.0 & 68.7 \\
    GeoCLIP~\cite{GeoCLIP} & 14.1 & 34.4 & 50.6 & 69.6 & 83.8 & & 9.5 & 19.3 & 32.6 & 55.0 & 74.6 \\
    PIGEOTTO~\cite{Pigeon} & 11.3 & 36.7 & 53.8 & 72.4 & 85.3 & & 10.4 & 23.7 & 40.6 & 62.2 & 77.7 \\
    \midrule
    \multicolumn{12}{l}{\textit{Generation-based Methods}} \\
    Qwen2.5-VL~\cite{Qwen2.5-VL} & 2.5 & 15.5 & 29.2 & 40.3 & 49.2 & & 3.9 & 23.7 & 44.5 & 61.5 & 74.9 \\
    Img2Loc~\cite{Img2Loc} & 7.9 & 23.3 & 29.9 & 40.1 & 51.1 & & 7.9 & 14.2 & 19.5 & 29.9 & 39.7 \\
    G3~\cite{G3} & 14.3 & 35.8 & 49.4 & 66.9 & 81.7 & & 23.9 & 33.6 & 44.0 & 60.6 & 75.9 \\
    GeoReasoner~\cite{GeoReasoner} & - & 26.9 & 36.6 & 52.3 & - & & - & - & - & - & - \\
    GaGA (Ours) & 11.7 & 33.0 & 48.0 & 67.1 & 82.1 & & 6.9 & 18.9 & 34.5 & 56.7 & 71.6 \\
    \textbf{GaGA (MP16-Pro)} & \textbf{15.0} & \textbf{37.1} & \textbf{49.5} & \textbf{67.3} & \textbf{82.4} & & \textbf{24.3} & \textbf{33.7} & 43.9 & \textbf{61.4} & \textbf{76.3} \\
    \bottomrule
    \end{tabular}}
\end{table}

\section{Hyper-parameter Settings}
\label{app:exp_setting}

Tables~\ref{tab:pretrain-settings} and~\ref{tab:finetune-settings} summarize the hyper-parameters used in the two training stages. Parameters not mentioned in Stage 2 are the same as those in Stage 1.

\begin{table}[ht]
    \centering
    \begin{minipage}[t]{0.48\linewidth}
        \centering
        \caption{Stage 1 settings.}
        \label{tab:pretrain-settings}
        \resizebox{\linewidth}{!}{%
            \begin{tabular}{ll}
                \toprule
                \textbf{Configuration} & \textbf{Value} \\
                \midrule
                Dataset & Meta Part of MG-Geo \\
                Training Epochs & 1 \\
                Total Batch Size & 16 \\
                Optimizer & AdamW \\
                LR & 2$\times$10$^{-4}$ \\
                LR Schedule & CosineAnnealing \\
                Weight Decay & 0 \\
                Warmup Ratio & 0.03 \\
                Adam Beta1 & 0.9 \\
                Adam Beta2 & 0.999 \\
                Image Resolution & 336$\times$336 \\
                Max Text Token Length & 1472 \\
                \bottomrule
            \end{tabular}}
    \end{minipage}
    \hfill
    \begin{minipage}[t]{0.48\linewidth}
        \centering
        \caption{Stage 2 settings.}
        \label{tab:finetune-settings}
        \resizebox{\linewidth}{!}{%
            \begin{tabular}{ll}
                \toprule
                \textbf{Configuration} & \textbf{Value} \\
                \midrule
                Dataset & 240K mixture from Meta, Clue, and Dialog Parts \\
                Training Epochs & 1 \\
                Total Batch Size & 16 \\
                Optimizer & AdamW \\
                LR & 2$\times$10$^{-5}$ \\
                Quantization Type & BitsAndBytesConfig \\
                Quantization Bits & 4-bit \\
                4-bit Quant Type & nf4 \\
                4-bit Compute Dtype & torch.float16 \\
                LoRA Alpha & 16 \\
                Low-Rank Matrix Rank & 64 \\
                LoRA Dropout & 0.05 \\
                \bottomrule
            \end{tabular}}
    \end{minipage}
\end{table}

\section{Prompts Employed in MG-Geo Dataset Generation}
\label{app:prompts}
To ensure question variety, we design multiple templates for each question type. These templates provide variation while maintaining focus on the geolocation task. The prompts below show the instruction templates used for clue extraction and dialogue generation.

\begin{mdframed}[backgroundcolor=black!5,linecolor=black!75,linewidth=0.5pt]
\textbf{Clue Part Prompt}\\[2pt]
(1) Analyze the given image for clues that help in geolocation and combine these clues to localize the image. Output the answer in JSON format.\\
(2) Can you identify the place where this image was taken? Analyze the street-view image from multiple angles to infer its geographic location and output the results and clues in JSON format.\\
(3) Where was this image taken? Analyze the image in conjunction with the geographic clues in the image. Output localization results and inference clues in JSON format.
\end{mdframed}

The \textit{CoT Deduction} prompt guides the model through reasoning and prediction, while the \textit{CoT Introspection} prompt asks the model to re-check evidence under a verified geographic anchor.

\begin{mdframed}[backgroundcolor=black!5,linecolor=black!75,linewidth=0.5pt]
\textbf{CoT Deduction Prompt}\\[2pt]
[Role Setting]\\
You are an excellent GeoGuessr player and questioner. The player deduces the location step by step from clues such as environment, climate, buildings, culture, and appearance, while the questioner guides deeper analysis to uncover more clues.\\
{[Reasoning QA]}\\
1. Based on the provided image, conduct THREE rounds of QAs (Q1A1, Q2A2, and Q3A3) between the questioner and the player.\\
2. Questions should be sufficiently challenging and closely related to the visual elements, but should NOT directly provide visual details to the player.\\
3. Only include questions that guide position prediction and require the player to use complex reasoning, world knowledge, and interpretive answers to gradually deduce the location. When answering complex questions, provide detailed reasoning steps for clarity and persuasiveness.\\
{[Coordinate Prediction]}\\
1. After the reasoning, the questioner should ask about the geographic coordinates and request an answer from the player, denoted as Q4A4.\\
2. Based on the previous rationale and analysis, the player makes the best prediction and briefly explains the choice. The player MUST provide reasonable coordinates regardless of uncertainty.\\
Please use Decimal Degrees for coordinates and STRICTLY follow this JSON format:\\
\{(latitude, longitude)\}
\end{mdframed}

\begin{mdframed}[backgroundcolor=black!5,linecolor=black!75,linewidth=0.5pt]
\textbf{CoT Introspection Prompt}\\[2pt]
[Verified Anchor]\\
The initial coordinate prediction is outside the metadata tolerance. The verified geographic coordinates are \{(X, Y)\}. Use them as an anchor for evidence re-checking, not as a shortcut answer.\\
{[Evidence Re-checking]}\\
1. Re-examine whether each previous clue is directly visible in the image.\\
2. Remove or revise any clue that is not visually grounded.\\
3. Identify overlooked high-specificity evidence, such as road markings, driving side, language, architecture, vegetation, or landmark-specific details.\\
4. Explain how the retained visual evidence supports the verified location and where the previous reasoning failed.\\
{[Request]}\\
1. Keep the original question structure Q1--Q4.\\
2. Use a reasoning tone, but do not introduce non-visible or unsupported clues.\\
3. Provide the final coordinates in A4 using Decimal Degrees and strictly follow this JSON format:\\
\{(latitude, longitude)\}
\end{mdframed}

\section{Cue Distribution in the Clue and Dialog Parts}
\label{app:cue_distribution}

\begin{figure}[!h]
    \centering
    \includegraphics[width=1\columnwidth]{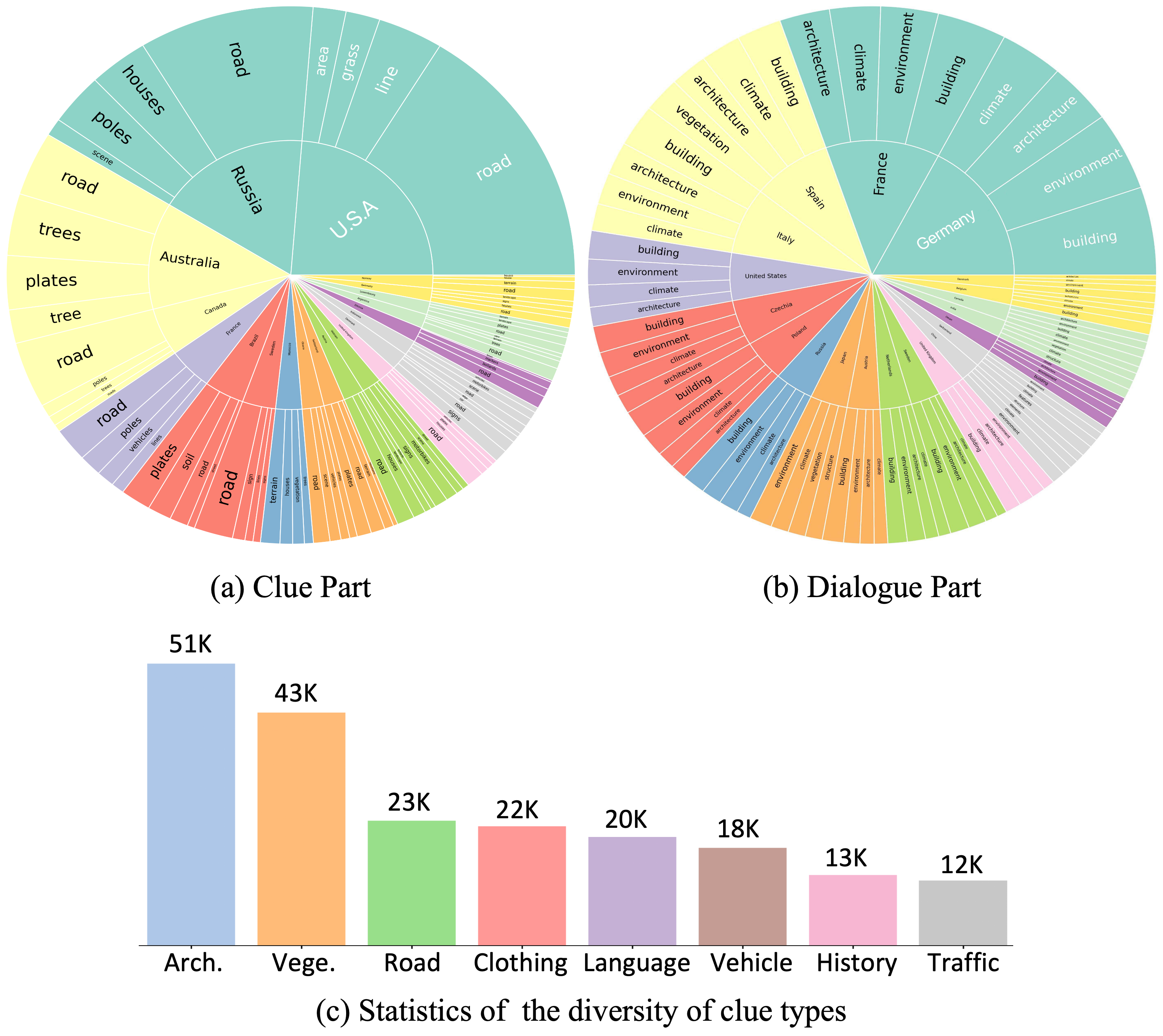}
    \caption{Detailed analysis of the Clue Part and Dialog Part.}
    \label{fig:statistic}
\end{figure}

We further analyze the cue distribution in the Clue and Dialog Parts. For the top-20 countries by sample frequency, we extract the four most frequent direct cue phrases. As shown in Fig.~\ref{fig:statistic}, the Clue Part covers more fine-grained geographic evidence, while the Dialog Part emphasizes conversational reasoning around buildings, environment, architecture, and climate. Across the eight cue categories, architectural features, vegetation, and road styles are the most frequent evidence types in street-view geolocation.

\section{Performances of Advanced MLLMs in Dialog}
\label{app:advanced_mllm_dialog}

Table~\ref{tab:advanced_vlm} illustrates the performance of various MLLMs under different interaction paradigms, comparing \textbf{Direct Inquiry} against \textbf{Tier-1 Focus-based Inquiry}. Comparing GaGA with its base model, LLaVA-Llama3, we observe that under Focus-based Inquiry, GaGA achieves precision gains at the region and city levels. The country-level decline under Tier-1 focus-based inquiry also reveals a limitation: non-discriminative focus questions can make the model over-attend to weak visual motifs, causing brittle reasoning under ambiguous cues. Tier-2 factual anchors mitigate this issue by converting broad visual attention into testable geographic constraints.

Furthermore, we evaluate the impact of this interactive design on other general-purpose MLLMs, including InternVL2~\cite{Internvl2} and Qwen-VL~\cite{Qwen-VL}. While Qwen-VL struggles under the standard Direct Inquiry setting, its country-level accuracy improves significantly when prompted with a Focus-based Inquiry. Similarly, through guided dialogue, InternVL2 uncovers actionable visual clues and improves across administrative levels. These results suggest that interactive reasoning can activate latent geographic knowledge and mitigate visual ambiguity across different MLLM architectures.

We employ two evaluation modes, \textbf{hierarchical (HIER)} and \textbf{direct (DIRE)}. HIER is primarily used for MLLMs that have not been finetuned on MG-Geo; in this mode, candidate administrative boundary names are provided at each level to constrain the output space.

\begin{table}[h]
\caption{Performance of advanced MLLMs with different types of prompt inputs.}
\centering
\resizebox{0.8\linewidth}{!}{%
    \begin{tabular}{l|c|c|c|ccc}
    \toprule
    \multirow{2}{*}{\textbf{Method}} & \multirow{2}{*}{\makecell[c]{\textbf{Evaluation} \\ \textbf{Mode}}} & \multirow{2}{*}{\textbf{Prompt}} & \multirow{2}{*}{\textbf{Recall}} & \multicolumn{3}{c}{\textbf{Admin-Level Accuracy}} \\
         & & & & Country & Region & City \\
    \midrule
    \multirow{3}{*}{GaGA} & \multirow{3}{*}{DIRE}
        &  Direct inquiry & 1 & 64.89 & 27.97 & 7.67 \\
        & & \multirow{2}{*}{Focus-based Inquiry}
        & \multirow{2}{*}{1}
        & 61.24 & 29.25 & 8.22\\
        & & & & \textcolor{s_blue}{-3.65} & \textcolor{darkgreen}{+1.28} & \textcolor{darkgreen}{+0.55} \\
    \midrule
    \multirow{3}{*}{InternVL2} & \multirow{3}{*}{HIER}
        &  Direct inquiry & 0.96 & 54.11 & 19.19 & 3.29 \\
        & & \multirow{2}{*}{Focus-based Inquiry}
         & \multirow{2}{*}{0.97}
          & 55.02 & 19.19 & 4.57\\
          & & & & \textcolor{darkgreen}{+0.91} & \textcolor{darkgreen}{0} & \textcolor{darkgreen}{+1.28} \\
    \midrule
    \multirow{3}{*}{Qwen-VL} & \multirow{3}{*}{HIER}
        &  Direct inquiry & 0.96 & 13.89 & 6.03 & 2.01 \\
        & & \multirow{2}{*}{Focus-based Inquiry}
         & \multirow{2}{*}{0.92}
          & 21.38 & 6.94 & 1.82\\
          & & & & \textcolor{darkgreen}{+7.49} & \textcolor{darkgreen}{+0.91} & \textcolor{s_blue}{-0.19} \\
    \midrule
    \multirow{3}{*}{LLaVA-LlaMA3} & \multirow{3}{*}{HIER}
        &  Direct inquiry & 0.99 & 2.92 & 0.54 & 0 \\
        & & \multirow{2}{*}{Focus-based Inquiry}
         & \multirow{2}{*}{0.99}
          & 4.38 & 0.54 & 0.05\\
          & & & & \textcolor{darkgreen}{+1.46} & \textcolor{darkgreen}{0} & \textcolor{darkgreen}{+0.05} \\
    \bottomrule
    \end{tabular}}
\label{tab:advanced_vlm}
\end{table}

\section{Ablation on Training Stages}
\label{app:training_stage_ablation}

As shown in Table~\ref{tab:model_comparison}, we evaluate the impact of the two training stages. GaGA-Stage1 achieves slightly higher static administrative-level accuracy, while GaGA-Stage2 introduces dialog-based reasoning and improves the model's ability to integrate user-provided constraints. This trade-off is consistent with the goal of building an interactive geolocation assistant rather than a purely one-shot classifier.

\begin{table}[h]
    \centering
    \caption{Impact of training stages on GaGA's performance.}
    \label{tab:model_comparison}
    \small
    \begin{tabular}{l|c|c|ccc}
    \specialrule{\heavyrulewidth}{0pt}{0pt}
    \multirow{2}{*}{\textbf{Method}} & \multirow{2}{*}{\makecell[c]{\textbf{Evaluation} \\ \textbf{Mode}}} & \multirow{2}{*}{\textbf{Recall}} & \multicolumn{3}{c}{\textbf{Admin-Level Accuracy (\%)}} \\
     & & & Country & Region & City \\
    \specialrule{\lightrulewidth}{0pt}{0pt}
    LLaVA-Llama3 & HIER & 0.99 & 1.76 & 0.26 & 0.02 \\
    GaGA-Stage1 & DIRE & 0.99 & \textbf{63.38} & \textbf{28.84} & \textbf{6.47} \\
    GaGA-Stage2 & DIRE & 1.00  & \underline{63.06} & \underline{27.95} & \underline{6.28}\\
    \specialrule{\heavyrulewidth}{0pt}{0pt}
    \end{tabular}
\end{table}

\section{Additional Qualitative Results}
\label{app:qualitative}

\begin{figure}[!t]
    \centering
    \includegraphics[width=1\columnwidth]{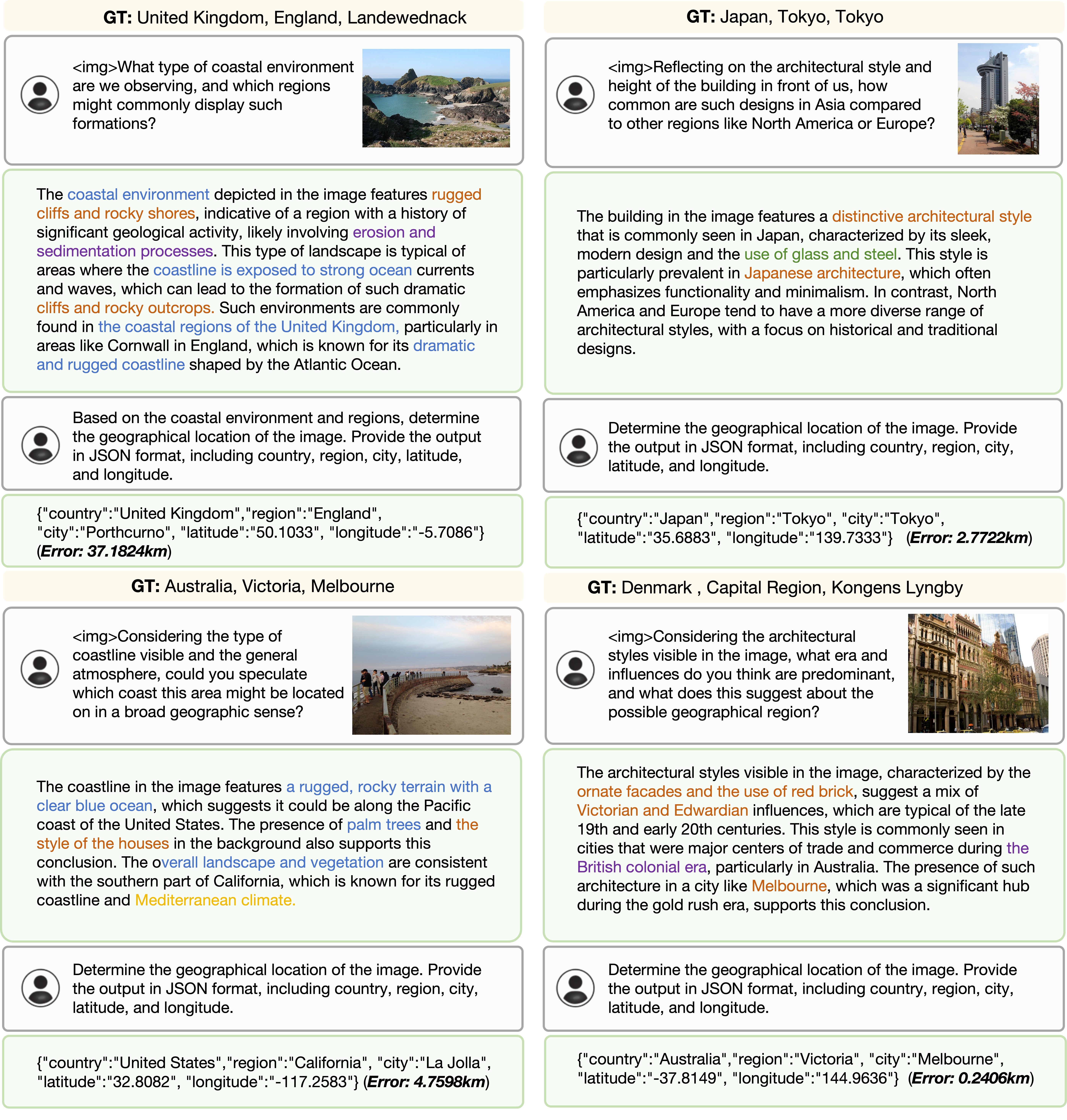}
    \caption{Additional qualitative results of GaGA.}
    \label{fig:vis_app_countryside}
\end{figure}

To further illustrate GaGA's localization behavior across diverse global scenarios, we present additional qualitative results in Fig.~\ref{fig:vis_app_countryside}.

\end{document}